\documentclass[sigconf]{acmart}
\usepackage[ruled,linesnumbered]{algorithm2e}
\usepackage{multirow}
\usepackage{color, colortbl}
\definecolor{LightCyan}{rgb}{0.88,1,1}
\AtBeginDocument{%
  \providecommand\BibTeX{{%
    \normalfont B\kern-0.5em{\scshape i\kern-0.25em b}\kern-0.8em\TeX}}}

\newcommand{\distillmethodone}{class-wise passing}
\newcommand{\distillmethodtwo}{pixel-wise passing}
\newcommand{\distillmethodthr}{instance-wise passing}
\newcommand{\distillmethodonecap}{Class-wise passing}
\newcommand{\distillmethodtwocap}{Pixel-wise passing}
\newcommand{\distillmethodthrcap}{Instance-wise passing}
\newcommand{\networkname}{SPNet}

\copyrightyear{2022} 
\acmYear{2022} 
\setcopyright{acmlicensed}\acmConference[MM '22]{Proceedings of the 30th ACM International Conference on Multimedia}{October 10--14, 2022}{Lisboa, Portugal}
\acmBooktitle{Proceedings of the 30th ACM International Conference on Multimedia (MM '22), October 10--14, 2022, Lisboa, Portugal}
\acmPrice{15.00}
\acmDOI{10.1145/3503161.3547891}
\acmISBN{978-1-4503-9203-7/22/10}

\begin{document}

\title{Paint and Distill: Boosting 3D Object Detection with Semantic Passing Network}


\author{Bo Ju}
\email{jubo@baidu.com}
\authornote{Equal Contribution.}
\orcid{0000-0002-8791-2656}
\affiliation{%
  \institution{Baidu Inc.}
  \city{Beijing}\country{China}
}

\author{Zhikang Zou}
\email{zouzhikang@baidu.com}
\authornotemark[1]
\orcid{0000-0003-3524-2942}
\affiliation{%
  \institution{Baidu Inc.}
  \city{Beijing}\country{China}
}

\author{Xiaoqing Ye}
\email{yexiaoqing@baidu.com}
\authornote{Corresponding author: Xiaoqing Ye.}
\orcid{0000-0003-3268-880X}
\affiliation{%
  \institution{Baidu Inc.}
  \city{Beijing}\country{China}
}

\author{Minyue Jiang}
\email{jiangminyue@baidu.com}
\orcid{0000-0002-4553-8643}
\affiliation{%
  \institution{Baidu Inc.}
  \city{Beijing}\country{China}
}

\author{Xiao Tan}
\email{tanxiao01@baidu.com}
\orcid{0000-0001-9162-8570}
\affiliation{%
  \institution{Baidu Inc.}
  \city{Beijing}\country{China}
}

\author{Errui Ding}
\email{dingerrui@baidu.com}
\orcid{0000-0002-1867-5378}
\affiliation{%
  \institution{Baidu Inc.}
  \city{Beijing}\country{China}
}

\author{Jingdong Wang}
\email{wangjingdong@baidu.com}
\orcid{0000-0002-4888-4445}
\affiliation{%
  \institution{Baidu Inc.}
  \city{Beijing}\country{China}
}

\begin{abstract}

3D object detection task from lidar or camera sensors is essential for autonomous driving. Pioneer attempts at multi-modality fusion complement the sparse lidar point clouds with rich semantic texture information from images at the cost of extra network designs and overhead. In this work, we propose a novel semantic passing framework, named SPNet, to boost the performance of existing lidar-based 3D detection models with the guidance of rich context painting, with no extra computation cost during inference. Our key design is to first exploit the potential instructive semantic knowledge within the ground-truth labels by training a semantic-painted teacher model and then guide the pure-lidar network to learn the semantic-painted representation via knowledge passing modules at different granularities: \distillmethodone{}, \distillmethodtwo{} and \distillmethodthr{}. Experimental results show that the proposed SPNet can seamlessly cooperate with most existing 3D detection frameworks with 1$\sim$5\% AP gain and even achieve new state-of-the-art 3D detection performance on the KITTI test benchmark. Code is available at: \url{https://github.com/jb892/SPNet}.

\begin{figure}[t]
\centering
 \includegraphics[width=0.48\textwidth]{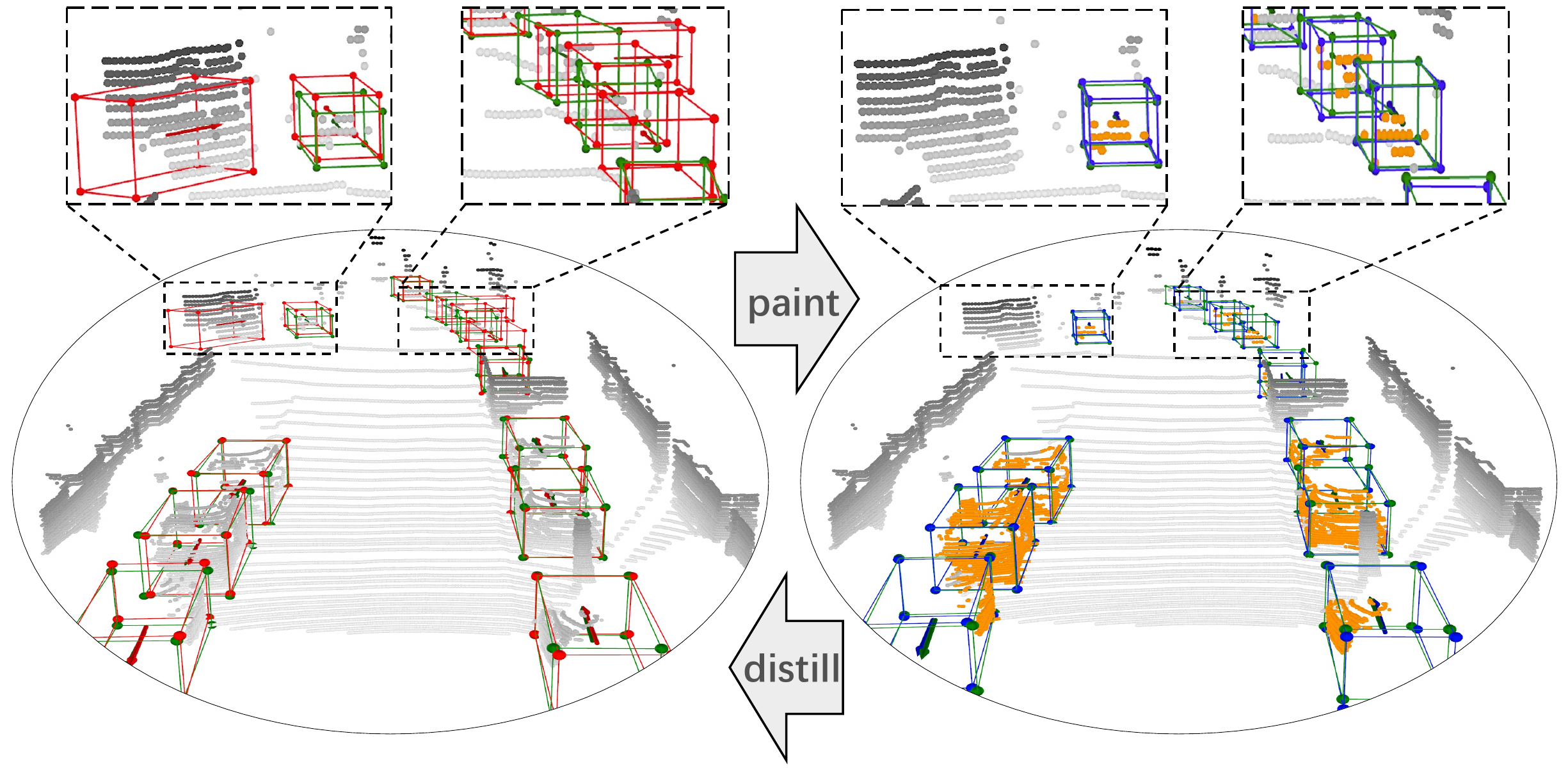}
 \caption{Left: the original point clouds and the red-colored prediction results. Right: the GT-painted point clouds in orange and the blue-colored corresponding prediction results. We use Pointpillars \cite{lang2019pointpillars} as the inference model and render the ground truth boxes in green. The results show that adding prior semantics to the raw point clouds can significantly improve the model ability on the hard examples (zoomed area in black dash line).}
 \label{fig:1}
\end{figure}

\end{abstract}




\maketitle
\section{Introduction}


3D object detection aims to estimate the informative 3D bounding boxes from lidar or camera sensors. It is a critical task and has received substantial attention for its wide applications in autonomous driving and robotics navigation. Recently emerged 3D object detection approaches mainly focus on learning discriminative 3D representations on the point clouds collected from LiDAR sensors, which provide accurate distance measurements and geometric cues for perceiving the surrounding environment. 3D point clouds are usually sparse, unordered, and unevenly distributed by nature. As a result, many approaches \cite{Shi_2020_CVPR,li2021anchor,li2021voxel,mao2021voxel} parse the sparse 3D point clouds into a compact representation and apply convolutional neural networks (CNN) to learn the structured features and explicitly explore the 3D shape information. Thanks to the robust representation ability of CNNs, remarkable progress has been achieved in promoting the performance on the existing mainstream benchmarks \cite{geiger2013vision,caesar2020nuscenes}. However, LiDAR points lack rich texture attributes and fine-grained shape knowledge of the objects, which limit the generalization performance of pure-lidar-based methods. Take the left part in Figure \ref{fig:1} as an example. We find that the LiDAR-based approach Pointpillars \cite{lang2019pointpillars} performs well on the easy and visible objects while the  localization capability deteriorates for the far-away or small objects, causing undesired false positives.

To alleviate the issues caused by the lack of semantic texture information in the point clouds, some approaches \cite{xu2018pointfusion,wang2021pointaugmenting,yin2021multimodal} try to explore the cross-modal fusion of RGB cameras and LiDAR sensors for complementarity to improve the performance. Compared to point clouds, RGB images contain informative colour and texture information but fail to recover the depth. The complementary nature of the two modalities makes them beneficial to each other and jointly leads to a win-win situation. Despite the promising results, these fusion-based methods suffer from three potential difficulties. 
First, LiDAR and camera image contain highly different data characteristics, making it  non-trivial to combine the features from these two misaligned views.
Second, the more complex multi-modal networks require extra computation head compared to single modality, which is undesirable for embedded systems. Third, the camera sensor is susceptible to the external influence, such as weather and light conditions, which may introduce undesired noise that hurts the final localization accuracy.
Given the problems mentioned above, we wonder if there is a way to dig into the underlying semantic information in the ground-truth label itself and eliminate the dependence on RGB images?


Towards this goal, we design a paint-to-distill framework in this paper, which explores the inherent semantics of objects to boost the performance of existing 3D detectors without extra inference cost.
Our work is inspired by the observations in the early stage fusion method PointPainting \cite{vora2020pointpainting}. It proposes a simple painting method that first assigns each point with the semantic label predicted from the semantic segmentation predictions on the 2D images and then trains the whole network using the painted point clouds as inputs. Although notable improvement is achieved on pedestrian and cyclist classes, it barely improves the performance of the commonly used car category. We claim that the dissatisfactory results are mainly bottle-necked by the imperfect segmentation outputs. Hence, we introduce GT-Painting, which paints the point clouds using the ground-truth bounding boxes and removes the dependence on RGB images. Since the ground truth bounding boxes can provide totally correct category semantic label, we can reach the potential performance upper bound of painting-based methods leveraging the ground truth semantic information rather than predicted segmentation clues. Experimentally, we demonstrate that using GT-Painting can improve the performance on the hard examples shown in the right part of Figure \ref{fig:1}. In addition, considering the moderate difficulty level, it achieves over 10\% AP gain of car category than the baseline model on the KITTI validation set.

Since the ground truth is not available during the inference stage, we design a novel distillation framework, named Semantic Passing Network (SPNet), to alleviate the labels dependence by distilling target-specific knowledge.
Firstly, we utilize the GT-painted point clouds for training a more robust teacher model. Then combine the ground truth bounding boxes and the deep features of various granularities in the pre-trained teacher to guide the student learning, which only accepts original point clouds as inputs. Since the features of the teacher model contain rich semantic information, this additional guidance can be complementary to the default box supervision in the training stage and discarded in the inference stage, which jointly contribute to the final performance. Specifically, we design three distillation schemes to pass the semantic information between models: (\romannumeral1) \distillmethodone{} module that models the group information in the 3D representations and provides guidance on clustering; (\romannumeral2) \distillmethodtwo{} module that aligns the feature maps in the BEV view; (\romannumeral3) \distillmethodthr{} module that closes the gap in the spatial distribution of the model outputs. The benefit of the above complementary information passing modules is that the teacher net and student net can be roundly aligned, which assists the student to distill much richer semantic information from the strong teacher. To verify the effectiveness of the proposed model, we plug our SPNet into three existing 3D object detectors and achieve consistent performance improvement. Note that our best model achieves new state-of-the-art 3D object detection performance on the KITTI benchmark. 

In conclusion, the major contributions are as follows:
\begin{itemize}
\item[$\bullet$] We propose a novel plug-and-play framework named Semantic Passing Network (SPNet), which uses the semantic information of objects to boost the performance of existing 3D object detectors at negligible cost.
\item[$\bullet$] We design three complementary information passing modules: \distillmethodone, \distillmethodtwo, \distillmethodthr, which provide auxiliary supervision signals which serve as strong guidance to transfer the rich semantics from the strong teacher to the student model. 
\item[$\bullet$] We achieve new state-of-the-art 3D object detection performance on the KITTI benchmark. Besides, we plug our model into three existing detectors and achieve significant improvements, with 2.84\%, 4.97\% and 1.05\% moderate AP increase on the KITTI benchmark.
\end{itemize}

\section{Related Work}

\subsection{3D Detection}
3D object detectors can be roughly divided into three streams: point-based, voxel-based and point-voxel-based, depending on how to transform point clouds into 3D representations for localizing objects. First, point-based detectors \cite{qi2018frustum,yang2019std,shi2019pointrcnn,shi2020point,pan20213d,chen2022sasa} operate directly on raw point clouds to generate 3D boxes. F-PointNet \cite{qi2018frustum} is a pioneering work that utilizes frustums for region-level features generation. PointRCNN \cite{shi2019pointrcnn} directly segments 3D point clouds to obtain foreground points and utilizes the segmentation features to refine the proposals. Pointformer \cite{pan20213d} proposes a local-global transformer to build globally-adaptive point representations by integrating local features with global features. SASA \cite{chen2022sasa} finds that the widely-used set abstraction (SA) is inefficient to describe scenes in the context of detection and thus incorporates point-wise semantic cues to focus on more informative foreground points. Second, voxel-based detectors \cite{zhou2018voxelnet,zhu2020ssn,deng2020voxel,li2021anchor,mao2021voxel} aim to transform the unstructured point clouds into regular grids over which conventional CNNs can be easily applied. VoxelNet \cite{zhou2018voxelnet} encodes the points to form a compact feature representation and produces detection results by using region proposal networks. Voxel R-CNN \cite{deng2020voxel} designs a new voxel RoI pooling operation to extract neighbouring voxel features, which are taken for further box refinement. VoTr \cite{mao2021voxel} further proposes the local self-attention module to replace the sparse convolution for voxel processing, which can be applied in most voxel-based detectors to boost the performance. Differently, CenterPoint \cite{Yin_2021_CVPR} proposes an anchor-free 3D detection framework, which uses a keypoint detector to regress centers of objects and other attributes. To integrate the best of the two representations, a series of point-voxel based methods \cite{Shi_2020_CVPR,li2021voxel,sheng2021improving,noh2021hvpr} have been proposed. The representative work PV-RCNN \cite{Shi_2020_CVPR} utilizes a novel voxel set abstraction module to encode representative scene features while proposing the RoI grid pooling to abstract proposal-specific features from the keypoints. These two innovative operations bring definite improvements. PV-RCNN++ \cite{shi2021pv} further replaces the RoI grid pooling with Vector-Pooling to efficiently collect the key point features from different orientations, which improves the performance with less resource consumption and faster running speed. To tackle the shape misses challenge caused by self-occlusion, BtcDet \cite{xu2021behind}  learns the object shape priors and estimates the complete object shapes that are partially curtained in point clouds. Thus, using the recovered shape miss can improve the 3D detection performance.


\begin{figure*}[t]
 \begin{center}
  \includegraphics[width=0.75\textwidth]{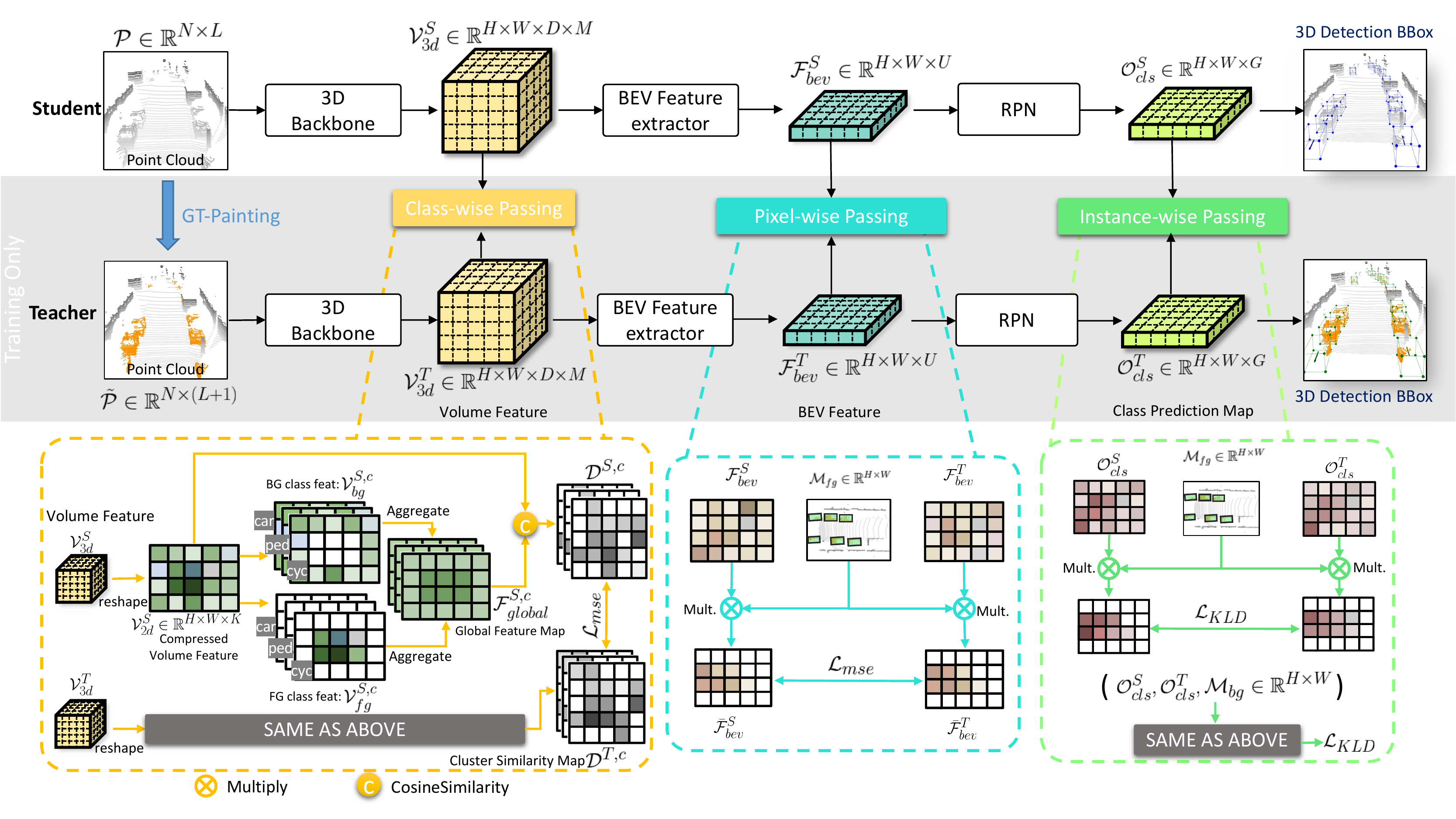}
 \end{center}
 \caption{
    The overview of our SPNet framework. It contains two branches: Student (Top) and Teacher (Bottom) with identical network structure. The student and teacher branches take raw point cloud and GT-painted point cloud into the network, respectively. The painted semantic representations learned by  the teacher network are passed to student network at different granularity levels $\{\mathcal{V}_{3d}$, $\mathcal{F}_{bev}$, $\mathcal{O}_{cls} \}$. Note that the teacher model is discarded in inference.
 }
 \label{spnet_framework}
\end{figure*}

\subsection{Knowledge Distillation}
Knowledge distillation has been extensively studied in recent years. The concept is popularized by Hinton et al. \cite{hinton2015distilling} that transfers the generalization ability from the teacher network to the student network through the soft targets. This procedure is realized by utilizing a distillation loss that considers both the ground truth and the prediction of the pre-trained teacher. Similar works can also be found in \cite{breiman1996born,fukuda2017efficient} but are designed mainly for classifiers. Since knowledge distillation can improve the performance of the student model while maintaining inference efficiency, it has been widely investigated in a variety of computer vision tasks, including object detection \cite{hao2020labelenc,guo2021distilling,dai2021general}, semantic segmentation \cite{liu2019structured,jiao2019geometry,shu2020channel}, object tracking \cite{liu2019teacher,dunnhofer2021weakly}, depth estimation \cite{hu2021boosting,chen2021revealing}. Due to the differences in backbones, it is not trivial to directly apply 2D methods to 3D tasks. As far as we know, there are few works \cite{wang2020multi,weidistill} that exploit knowledge distillation in LiDAR-based 3D object detection. Wang et al. \cite{wang2020multi} first train a teacher detection model on dense point clouds generated from multiple frames and then transfer the knowledge in the pre-trained teacher to the student network, which accepts sparse point clouds as inputs. Wei et al. \cite{weidistill} propose the LiDAR distillation method to mitigate the beam-induced domain shift induced by different LiDAR beams. Using extra semantic information as guidance, these methods tend to solve the serious sparseness problem in the point clouds. This paper focuses on exploiting the rich semantic information existing in the ground truth labels, which can boost the performance while maintaining the same setting as mainstream 3D object detectors.

\section{Methodology}

This paper presents a novel Semantic Passing Network (SPNet) for 3D object detection, depicted in Figure \ref{spnet_framework}. The core idea of this work is to make full use of the semantic information hidden in objects and map the semantics into auxiliary supervision signals to pass the instructive knowledge, which can boost the performance of existing 3D object detectors. Given original point clouds, we first use GT-painting to add the semantic class indicator on the point clouds and then use our SPNet to transfer the informative knowledge from the painted point clouds to the original point clouds. 

In the following section, we shall introduce the GT-Painting method in Sec. \ref{method_1}. Then, we give a thorough introduction about the proposed SPNet framework architecture in Sec. \ref{method_2}. In the last Sec. \ref{method_3}, we further explain the training losses of SPNet.

\subsection{GT-Painting}
\label{method_1}

Although LiDAR can provide depth information close to linearity error for accurate localization, it is limited by the lack of texture information. To resolve this issue, PointPainting \cite{vora2020pointpainting} proposes to leverage the rich semantics contained in RGB images to consolidate point clouds. Specifically, it appends the segmentation scores from per pixel predictions of the RGB image to the last dimension of the corresponding point cloud. As a result, the LiDAR-based 3D detectors benefit from the camera modality's rich semantic cues. However, through our observation, the performance of such a painting method is largely dependent on the accuracy of the 2D semantic segmentation result. Experiments show that the painting method could bring 2\% mAP performance improvement if we use a robust 2D semantic segmentation model like HRNet-OCR \cite{YuanCW19}. The improvements vanish if we adopt a weaker segmentation model like HRNet \cite{SunXLW19}. So, we design the GT-Painting to find the upper bound performance of the PointPainting method by directly using ground-truth boxes to provide the best semantic indicator and paint the point cloud. In particular, we define each raw input point cloud as $\mathcal{P} \in \mathbb{R}^{N\times L}$ and each point as $\{p_i = (x_i, y_i, z_i, r_i), i = 1,...N\}$, where $x_i, y_i, z_i$ represent the spatial location of each point, $r_i$ is the LiDAR reflectance intensity, and $N$ is the number of total points. We generate the semantic class indicator $\mathcal{I} \in \mathbb{R}^{N\times 1}$ by examining if the points are located inside any of the bounding boxes. Inner points with a specific class are marked as the corresponding numerical values and denoted as foreground. Otherwise, the points are marked as background with numerical value zero. Finally, we concatenate the raw point cloud $\mathcal{P}$ with the semantic binary indicator $\mathcal{I}$ in the channel dimension to get the painted point clouds $\tilde{\mathcal{P}}$. We feed these augmented point clouds to the baseline model PointPillars \cite{lang2019pointpillars} for training and achieve about 10\% moderate AP increase on the KITTI validation set. For the ground truth bounding boxes are not available in the inference stage, we design SPNet to alleviate the labels dependence while maintaining the privilege of exploiting rich semantic information.

\subsection{SPNet Architecture}
\label{method_2}
\noindent\textbf{Overview} The overview of the proposed {\networkname} framework structure is shown in Figure \ref{spnet_framework}. It follows a teacher-student paradigm, where the two models share the same structure. Since our network can be applied to most mainstream 3D detection models, we present our work using a general 3D detection framework rather than targeting a specific network. The model consists of three components: (\romannumeral1) a sparse convolution network (3D backbone) that transforms the input point clouds into 3D features $\mathcal{V}_{3d}$ for characterizing local 3D shape information; (\romannumeral2) an hourglass convolution network (BEV feature extractor) that flattens the 3D features $\mathcal{V}_{3d}$ into 2D BEV view and extract BEV representations $\mathcal{F}_{bev}$ full of the context information; (\romannumeral3) a region proposal network (RPN) that fuses high level abstract features $\mathcal{F}_{bev}$ to predict the category of objects $\mathcal{O}_{cls}$ and regress 3D bounding boxes $\mathcal{O}_{reg}$ simultaneously. In the first step, we use GT-Painting to decorate an input point cloud with ground truth labels and train the teacher model using painted point clouds as inputs. In the second step, the teacher model is initialized with pre-trained weights obtained earlier, while the student is initialized with random parameters. Finally, we use the features of the teacher model as additional supervision signals, which joint with the ground truth bounding boxes to supervise the student learning during the training time. In order to promote the sufficient delivery of semantic information between the two models, we design three complementary distillation modules: \distillmethodone{} module, \distillmethodtwo{} module and \distillmethodthr{} module. These modules are applied to the different stages in the student model to align the differences between the two models from different granularities.

\noindent\textbf{Class-wise Passing Module} 3D detection aims to estimate the 3D bounding box of the objects, whose essence is to perceive the clustering information of the point cloud and group points of the same type together. Such group-related information is often reflected in the feature distribution, so the difference in the clustering representation ability of different models is actually due to the feature distribution variance. To eliminate this variation, the straightforward way is to directly add a penalty term between the features of the two models to reduce the distance of the feature space. However, we claim that the features without explicit modelling will push the student model to mimic the teacher from a pixel-level perspective while losing global semantic information, which leads to the student being trapped in a local minimum. In fact, each point in the clustering has a strong semantic connection with the class center for a certain class of objects, which can be fully utilized to better reduce the feature distribution variation. Hence, we propose a class-wise passing module to transfer relation semantics from the teacher to the student model by modelling high-order statistics. Specifically, given the 3D features $\mathcal{V}^{T}_{3d}$, $\mathcal{V}^{S}_{3d} \in \mathbb{R}^{H\times W\times D\times M}$ taken from the 3D backbone, we first compress them along the gravitational dimension $D$ to get the BEV shape features $\mathcal{V}^{T}_{2d}$, $\mathcal{V}^{S}_{2d} \in \mathbb{R}^{H\times W\times K}$, where $K = D\times M$. BEV perspective provides dense features with richer spatial information while preserving 3D inherent structural information, which allows us to better compute constraint relations on the feature maps. In order to distinguish the objects of different classes, we compute the binary foreground mask $\mathcal{M}^c_{fg}\in \mathbb{R}^{H\times W}\text{ and background mask } \mathcal{M}^c_{bg}\in \mathbb{R}^{H\times W}$ for each class using the ground truth bounding boxes, where $c$ represents the specific object class. Next, we calculate the feature center of each class, which is realized by masking the feature maps of the same class and applying the average pooling to aggregate the features along the spatial dimension.
\begin{equation}
\mathcal{F}^{S,c}_{center} = \frac{\sum_{i=1}^ {H \times W} (\mathcal{M}^c_{fg} \cdot \mathcal{V}^S_{2d})}{\sum_{i=1}^ {H \times W} \mathcal{M}^c_{fg}}
\end{equation}
Then the center vector is unsqueezed to the same resolution as the mask, represented as $\tilde{\mathcal{F}}^{S,c}_{center}$. We combine the center map and the background using the mask as guidance to capture long-range global context information, which is:
\begin{equation}
\begin{aligned}
\mathcal{F}^{S,c}_{global} &= \mathcal{V}^{S, c}_{bg} + \mathcal{V}^{S, c}_{fg} \\
&                           = \mathcal{M}^c_{bg} \cdot \mathcal{V}^S_{2d} + \mathcal{M}^c_{fg}\cdot \tilde{\mathcal{F}}^{S,c}_{center}
 \end{aligned}
\end{equation}
In the same way, we can get the center vector $\mathcal{F}^{T,c}_{center}$ and further obtain the global feature map $\mathcal{F}^{T,c}_{global}$ for the teacher. To model the inner-class feature constraints, we compute the cluster similarity maps $\mathcal{D}^{T,c}, \mathcal{D}^{S,c}$ by establishing cosine similarity matrix between the global feature maps and the original BEV features for both teacher and student features.
\begin{equation}
\begin{aligned}
\mathcal{D}^{T, c} = \frac{\mathcal{V}^T_{2d}\cdot \mathcal{F}^{T,c}_{global}}{max\left ( \left \| \mathcal{V}^T_{2d}\right \|_2\cdot \left \| \mathcal{F}^{T,c}_{global} \right \|_2, \epsilon \right )} \\
\mathcal{D}^{S, c} = \frac{\mathcal{V}^S_{2d}\cdot \mathcal{F}^{S,c}_{global}}{max\left ( \left \|  \mathcal{V}^S_{2d}\right \|_2\cdot \left \| \mathcal{F}^{S,c}_{global} \right \|_2, \epsilon\right )}
\end{aligned}
\end{equation}
where the $\epsilon$ value is 1e-6 to prevent zero division. Since the similarity matrix contains rich semantics about the group relations, we calculate the {\distillmethodone} loss $\mathcal{L}_c$ between cosine similarity matrix of teacher and student, which is:
\begin{equation}
\mathcal{L}_c = \frac{1}{H\times W}\sum_{i=1}^ {H \times W}\sum^{C}_{c}\left \| \mathcal{D}^{T, c} - \mathcal{D}^{S, c} \right \|_2
 \label{eq_loss_group}
\end{equation}
where $C$ represents the total categories used in the training.

\noindent\textbf{Pixel-wise Passing Module} 
Spatial information is a key component of the decisive factors for the accurate localization. Thus, we design a pixel level alignment method, named pixel-wise passing module, to align structured information among spatial locations between the teacher and student models. In specific, we take the BEV feature map $\mathcal{F}_{bev} \in \mathbb{R}^{H\times W \times U}$ produced from the compact BEV feature extractor as input where $U$ is the feature channel of $\mathcal{F}_{bev}$. By minimizing the L2 distance between two feature maps, the high-level instructive information contained in the deep features flow from teacher to student spatially. However, the background regions often occupy a significant amount of contribution in feature loss, which is unnecessary and make the {\distillmethodtwo} module less effective. To mitigate the effects of such noise, we 
produce the foreground binary feature mask $\mathcal{M}_{fg} \in \mathbb{R}^{H\times W}$ using the labels, which makess the loss calculation only focuses on foreground area. The loss function can be formulated as:
\begin{equation}
\mathcal{L}_f = \frac{\mathcal{M}_{fg} \cdot \left \| \mathcal{F}^T_{bev} - \mathcal{F}^S_{bev} \right \|_2}{\sum_{i=1}^ {H \times W} \mathcal{M}_{fg}} 
\label{eq_loss_feature}
\end{equation}

\noindent\textbf{Instance-wise Passing Module} Prediction differences carry vital information on why students lag behind teachers, which can further complement the guidance of class-wise and pixel-wise imitation. Many pioneer works \cite{wang2019distilling,guo2021distilling,zhou2021sgm3d} have proven that labels generated from a better domain can provide additional knowledge guidance at a fine-grained level of supervision instead of only using hard labels as direct supervision. Hence, we design an {\distillmethodthr} module that helps to increase the consistency between teacher and student branches in prediction level. The input of the {\distillmethodthr} module is the class prediction score maps $\mathcal{O}^{T}_{cls}, \mathcal{O}^{S}_{cls} \in \mathbb{R}^{H\times W \times G}$ of the teacher and student model, where $G$ is the number of anchors at each prediction grid. First, we adopt the masked Kullback-Leibler (KL) divergence loss to compute the distance between teacher and student prediction for both foreground and background. Then, we re-weight the foreground and background loss by using the empirical value $\lambda_{fg}$ and $\lambda_{bg}$ respectively. Finally, we add up two losses to get the final loss for the {\distillmethodthr} module.

\begin{equation}
\begin{split}
\mathcal{L}_{p} = & \lambda_{fg} \cdot KLD \left ( \mathcal{O}^{T}_{cls}, \mathcal{O}^{S}_{cls}, \mathcal{M}_{fg}\right ) + \\
& \lambda_{bg} \cdot KLD\left ( \mathcal{O}^{T}_{cls}, \mathcal{O}^{S}_{cls}, \mathcal{M}_{bg}\right )
 \end{split}
 \label{eq_prediction_wise_loss}
\end{equation}
where $\mathcal{M}_{bg}$ and $\mathcal{M}_{fg}$ represent the mask of the background and foreground, respectively. The masked K-L divergence loss can be described as: 

\begin{equation}
KLD\left ( \mathcal{F}^T, \mathcal{F}^S, \mathcal{M}\right ) = \frac{\sum (\mathcal{M}\cdot\mathcal{F}^T \cdot (log\mathcal{F}^T - log\mathcal{F}^S))}{max\left ( \sum \mathcal{M}, \epsilon \right )}
 \label{eq_kld_loss}
\end{equation}

\subsection{Overall loss function} 
\label{method_3}
We strictly follow the detection losses used in the adopted baselines. Take PointPillars \cite{lang2019pointpillars} as a representative, we use focal loss $\mathcal{L}_{cls}$ for object classification, a softmax classification loss $\mathcal{L}_{dir}$ on the discretized directions and smooth L1 loss $\mathcal{L}_{loc}$ for the localization regression residuals between ground truth and anchors. More details can be found in complementary details. The total 3D detection loss is therefore:
\begin{equation}
\mathcal{L}_{3d} = \lambda_{cls}\cdot\mathcal{L}_{cls} + \lambda_{dir}\cdot\mathcal{L}_{dir} + \lambda_{loc}\cdot\mathcal{L}_{loc}
\end{equation}
Besides, we combine the distill losses from three information passing modules with the 3D detection losses to optimize the student model jointly. The total loss can be defined as:
\begin{equation}
\mathcal{L}_{total} = \mathcal{L}_{3d} + \lambda_c\cdot\mathcal{L}_{c} + \lambda_f\cdot\mathcal{L}_{f} + \lambda_p\cdot\mathcal{L}_{p}
 \label{eq_loss_total}
\end{equation}

\section{Experiments}




\subsection{Setup}
\textbf{Dataset} We evaluate our method on the challenging KITTI dataset \cite{geiger2013vision}, which consists of 7481 training samples and 7518 testing samples collected from autonomous driving scenes. Each sample provides both the point cloud and the camera image, while our method only uses the point cloud as input. Following the frequently used train/val split mentioned in the previous work \cite{chen2017multi}, we divide the training samples into train split of 3712 samples and val split of 3769 samples, respectively. We conduct ablation studies based on this split and report test results trained with all 7481 samples on the KITTI benchmark.

\noindent\textbf{Metrics} The KITTI dataset evaluates both the average precision (AP) performance of bird’s eye view (BEV) detection and 3D object detection. The labels are divided into three difficulty levels: easy, moderate and hard based on the basis of object size, occlusion and truncation levels. We report the 40 recall positions-based metric (R40) on the official KITTI test server and the 11 recall positions-based metric (R11) on the validation set for a fair comparison with previous works. We mainly focus on the Car category while also presenting Pedestrian and Cyclist performances. The rotated IoU threshold for three categories are 0.7, 0.7 and 0.5, respectively.

\begin{table}[t]
 \caption{Performance comparison on the KITTI test benchmark for the car category with 40 recall points. * means results are reproduced by the public OpenPCDet \cite{od2020openpcdet}.}
 \begin{tabular}{c|c|ccc}
  \toprule
  \multirow{2}{*}{Method} & \multirow{2}{*}{Reference} & \multicolumn{3}{c}{3D Detection (\%)} \\
                       &             & Mod.  & Easy & Hard \\
 \hline
 \hline
 \textbf{\textit{LiDAR + RGB:}} &  & & & \\
  MV3D \cite{chen2017multi}                   & CVPR 2017 & 63.63 & 74.97         & 54.00               \\
  F-PointNet \cite{qi2018frustum}              & CVPR 2018 & 69.79 & 82.19         & 60.59               \\
  Painting \cite{vora2020pointpainting}        & CVPR 2020 & 71.70 & 82.11         & 67.08               \\
  F-ConvNet\cite{wang2019frustum}              & IROS 2019 & 76.39 & 87.36         & 66.69               \\
  MMF \cite{liang2019multi}                    & CVPR 2019 & 77.43 & 88.40         & 70.22               \\
  3D-CVF \cite{yoo20203d}                      & ECCV 2020 & 80.05 & 89.20         & 73.11               \\
  CLOCs \cite{pang2020clocs}                   & IROS 2020 & 80.67 & 88.94         & 77.15               \\
  \hline
  \hline
  \textbf{\textit{LiDAR only:}} &  & & & \\
  VoxelNet \cite{zhou2018voxelnet}           & CVPR 2018    & 64.17  & 77.82           & 57.51               \\
  SECOND    \cite{yan2018second}             & Sensor 2018  & 72.55  & 83.34           & 65.82               \\
  PointRCNN    \cite{shi2019pointrcnn}       & CVPR 2019    & 75.64  & 86.96           & 70.70               \\
  Part-$A^2$ \cite{shi2020points}            & PAMI 2020    & 78.49  & 87.81           & 73.51               \\
  3DSSD \cite{yang20203dssd}                 & CVPR 2020    & 79.57  & 88.36           & 74.55               \\
  STD \cite{yang2019std}                     & ICCV 2019    & 79.71  & 87.95           & 75.09               \\
  IA-SSD        \cite{zhang2022IASSD}       & CVPR 2022    & 80.32  & 88.87           & 75.10               \\
  PV-RCNN     \cite{Shi_2020_CVPR}          & CVPR 2020    & 81.43  & 90.25           & 76.82               \\
  Voxel-Point \cite{li2021voxel} & MM 2021 & 81.58 & 88.53 & 77.37 \\
  Voxel-RCNN     \cite{deng2020voxel}        & AAAI 2021    & 81.62  & \underline{90.90}  & 77.06               \\
  CT3D \cite{sheng2021improving}            & ICCV 2021    & 81.77  & 87.83           & 77.16               \\
  VoxSeT \cite{he2022voxset}                 & CVPR 2022    & \underline{82.06}  & 88.53           & \underline{77.46}               \\
  \hline
  \hline
  PointPillars \cite{lang2019pointpillars}  & CVPR 2019 & 74.99  & 79.05          & 68.30          \\
  \textbf{SPNet-P}             & - & \textbf{77.83} & \textbf{85.84} & \textbf{72.84} \\
  \rowcolor{LightCyan}
  \textbf{\textit{Improvement}}         & - & \textit{+2.84} & \textit{+6.79} & \textit{+4.54} \\
  \hline
  *CenterPoint \cite{Yin_2021_CVPR}           & CVPR 2021 & 73.96  &   81.17        & 69.48          \\
  \textbf{SPNet-V}             & - & \textbf{78.93} & \textbf{86.87} & \textbf{73.64} \\
  \rowcolor{LightCyan}
  \textbf{\textit{Improvement}}         & - & \textit{+4.97} & \textit{+5.70} & \textit{+4.16} \\
  \hline
  *PVRCNN++  \cite{shi2021pv}            & Arxiv 2021 & 81.06 & 86.95           & 76.60          \\
  \textbf{SPNet-PV}       & - & \textbf{82.11} & \textbf{88.53} & \textbf{77.41} \\
  \rowcolor{LightCyan}
  \textbf{\textit{Improvement}}         & - & \textit{+1.05} & \textit{+1.58} & \textit{+0.81} \\
  \bottomrule
 \end{tabular}
 \label{kitti_test}
\end{table}

\noindent\textbf{Implementation Details} Our plug-and-play SPNet has three versions: (a) SPNet-P: is build upon the pillar-based method PointPillars \cite{lang2019pointpillars}, which is also anchor-based. (c) SPNet-V: is build upon the voxel-based method CenterPoint \cite{Yin_2021_CVPR}, which is also anchor-free. (c) SPNet-PV: is build upon the point-voxel-based method PV-RCNN++ \cite{shi2021pv}, which leads the top performance on the KITTI benchmark. Since our experimental settings are strictly consistent with the baseline we used, we only introduce the details of SPNet-P here. In the distillation stage, the loss weights for {\distillmethodone} $\lambda_c$, {\distillmethodtwo} $\lambda_f$, {\distillmethodthr} $\lambda_p$ are set to 0.1, 10 and 10, respectively. In \distillmethodthr{} module, the $\lambda_{fg}$ and $\lambda_{bg}$ are set to 2 and 0.1, respectively. In the 3D detection loss, the $\lambda_{cls}$, $\lambda_{dir}$ and $\lambda_{loc}$ are set to 1, 0.2 and 2, respectively. We train the student network from scratch in an end-to-end manner with ADAM optimizer for 80 epochs on 4 V100 GPUs. The learning rate is set to 0.003 with a poly rate using power as 0.9 and a weight decay of 0.9. The batch size per GPU is set to 4. For KITTI dataset, the x, y, z range of [(0, 69.12), (-39.68, 39.68), (-3, 1)] meters respectively, which is voxelized with the voxel size (0.16m, 0.16m, 4m) in each axis. During training, we adopt the common used data augmentation strategies for 3D object detection, including random world flip along the X axis, random world rotation around the Z axis with a random angle sampled from [-$\pi/4$, $\pi/4$], random world scaling with a random scaling ratio sampled from [0.95, 1.05]. Besides, we conduct the ground-truth sampling augmentation used in SECOND \cite{yan2018second} to crop several new boxes from ground truth and corresponding points in other frames and paste them in the training scenes.

\begin{table}[t]
 \caption{Performance comparison on the KITTI validation set for the car category with 11 recall points. * means results are reproduced by the public OpenPCDet \cite{od2020openpcdet}.}
 \centering
 \begin{tabular}{c|c|ccc}
  \toprule
  \multirow{2}{*}{Method}  & \multirow{2}{*}{Reference} & \multicolumn{3}{c}{3D Detection(\%)} \\
    &  & Mod.          & Easy            & Hard        \\
\hline
\hline
 \textbf{\textit{LiDAR + RGB:}} &  & & & \\
  MV3D \cite{chen2017multi} & CVPR 2017 & 62.28 & 71.29 & 56.56 \\
  F-PointNet \cite{qi2018frustum}  & CVPR 2018 & 70.92 & 83.76 & 63.65  \\
  3D-CVF \cite{yoo20203d}   & ECCV 2020 &  79.88 &  89.67 & 78.47  \\
\hline
\hline
 \textbf{\textit{LiDAR only:}} &  & & & \\
   SECOND \cite{yan2018second} & Sensor 2018 & 76.48 & 87.43 & 69.10 \\
   TANet \cite{liu2020tanet} & AAAI 2020 & 76.64 & 87.52 & 73.86 \\
   PointRCNN \cite{shi2019pointrcnn} & CVPR 2019 & 78.63 & 88.88 & 77.38 \\
   3DSSD \cite{yang20203dssd} & CVPR 2020 & 79.45 & 89.71 & 78.67 \\
   IA-SSD \cite{zhang2022IASSD} & CVPR 2022 & 79.57 & - & - \\
   PV-RCNN \cite{Shi_2020_CVPR} & CVPR 2020 &  83.69 & 89.35 & 78.70 \\
  VoTR-TSD \cite{mao2021voxel} & ICCV 2021  & 84.04 & 89.04 & 78.68 \\
  Pyramid-PV \cite{mao2021pyramid} & ICCV 2021 &
  84.38 & 89.37 & 78.84 \\
  Voxel-RCNN \cite{deng2020voxel}   & AAAI 2021 &  \underline{84.52} & 89.41 & \underline{78.93} \\
\hline
\hline
  PointPillars \cite{lang2019pointpillars}            & CVPR 2019 & 77.31 & 87.29 & 75.55  \\
  \textbf{SPNet-P}    & -   & \textbf{78.67} & \textbf{88.71} & \textbf{77.29}  \\
  \rowcolor{LightCyan}
  \textbf{\textit{Improvement}}         & - & \textit{+1.36} & \textit{+1.42} & \textit{+1.74} \\
  \hline
  *CenterPoint \cite{Yin_2021_CVPR}   &  CVPR 2021 &         76.98 & 86.76 & 74.54  \\
  \textbf{SPNet-V}      & -  & \textbf{78.32} & \textbf{87.88} & \textbf{77.24} \\
  \rowcolor{LightCyan}
  \textbf{\textit{Improvement}}         & - & \textit{+1.34} & \textit{+1.12} & \textit{+2.70} \\
  \hline
  *PVRCNN++ \cite{shi2021pv}               & Arxiv 2021    & 83.84 & \textbf{89.34} & 78.81 \\
  \textbf{SPNet-PV}    & -           & \textbf{84.92} & 89.26 & \textbf{78.92} \\
  \rowcolor{LightCyan}
  \textbf{\textit{Improvement}}         & - & \textit{+1.08} & \textit{-0.08} & \textit{+0.11} \\
  \bottomrule
 \end{tabular}
 \label{kitti_val}
\end{table}
\begin{table}[t]
 \caption{Performance comparison on the KITTI validation set for the pedestrian and cyclist category with 11 recall points. The results are evaluated by average precision at IoU = 0.5.}
 \centering
 \resizebox{\columnwidth}{!}{
 \begin{tabular}{c|ccc|ccc}
  \toprule
  \multirow{2}{*}{Method}                & \multicolumn{3}{c|}{Pedestrian} & \multicolumn{3}{c}{Cyclist}   \\
                                           & Mod.           &  Easy          & Hard     & Mod.           &  Easy          & Hard       \\
\hline
\hline
Second \cite{yan2018second} & 51.84 & 57.02  & 47.38 & 65.21 & 82.00  & 61.35 \\
PointRCNN \cite{shi2019pointrcnn} & 58.32 & 63.29 & 51.59 & \underline{66.67} & 83.68 & \underline{61.92} \\
VoxelNet \cite{zhou2018voxelnet} & \underline{59.84} & \underline{67.79} & \underline{54.38} & 64.89 & \underline{84.92} & 58.59 \\  
\hline
\hline
  PointPillars \cite{lang2019pointpillars}                           & 54.98 & 59.42 & 49.86 & 65.19 & 81.73 & 62.20 \\
  SPNet-P                      & \textbf{56.46} & \textbf{61.07} & \textbf{51.19} & \textbf{68.03} & \textbf{82.10} & \textbf{63.00} \\
  \rowcolor{LightCyan}
  \textbf{\textit{Improvement}}         & \textit{+1.48} & \textit{+1.65} & \textit{+1.33} & \textit{+2.84} & \textit{+0.37} & \textit{+0.80}\\
  \bottomrule
 \end{tabular}}
 \label{kitti_ped_cyc}
\end{table}
\begin{table}[t]
 \caption{Effectiveness of each individual component in the SPNet for the car class at IOU=0.7 (R11).}
 \centering
 \begin{tabular}{c|ccc|ccc}
  \toprule
   \multirow{2}{*}{Group} & Class           & Feature      & Prediction      & \multicolumn{3}{c}{3D Detection (\%)}     \\
   & wise            & wise     & wise           & Mod.                & Easy                 & Hard    \\
  \hline
  \uppercase\expandafter{\romannumeral1} & $\times$         & $\times$ & $\times$         & 77.31               & 87.29                & 75.55             \\
  \uppercase\expandafter{\romannumeral2} & $\checkmark$     & $\times$ & $\times$         & 78.23    & 88.55    & 76.82  \\
  \uppercase\expandafter{\romannumeral3} & $\times$         & $\checkmark$ & $\times$     & 78.47   & 88.29     & 77.12  \\
  \uppercase\expandafter{\romannumeral4} & $\times$         & $\times$ & $\checkmark$     & 77.74   & 86.99    & 76.47  \\
   \uppercase\expandafter{\romannumeral5} & $\checkmark$     & $\checkmark$ & $\checkmark$ & \textbf{78.67 } & \textbf{88.71} & \textbf{77.29}\\
  \bottomrule
 \end{tabular}
 \label{ablation}
\end{table}

\subsection{Comparison to State-of-the-Art}
\textbf{Results on the KITTI test set} We first compare our SPNet with other state-of-the-art approaches on the KITTI test benchmark for the commonly used car category using the default IOU of 0.7. As shown in Table \ref{kitti_test}, our method of different versions SPNet-P, SPNet-V and SPNet-PV outperform the baseline methods consistently with large margins of 2.84\%, 4.97\% and 1.05\% mAP increase on the moderate level. Note that our SPNet-PV achieves the new state-of-the-art performance among all competitors with mAP of 82.11\%. In specific, compared with the strong competitor CLOCs \cite{pang2020clocs}, which combines the RGB images and the point clouds to boost the performance, our SPNet-PV achieves a significant improvement of 1.44\% on moderate AP. It is worth noting that CLOCs \cite{pang2020clocs} 
builds its best model based on the baseline PV-RCNN \cite{Shi_2020_CVPR} but still gets inferior results to us. Even compared with the very recent public method VoxSet \cite{he2022voxset}, which leads the top performance on the KITTI benchmark, our method can still achieve better performance. Since the proposed SPNet needs no extra computational cost during the inference time, our different variants of SPNet can achieve the same inference speed as the baseline models.

\noindent\textbf{Results on the KITTI val set} We evaluate the proposed framework of three versions with several state-of-the-art methods on the KITTI validation set. As shown in Table \ref{kitti_val}, the results are calculated by recall 11 positions with the IoU threshold 0.7 for a fair comparison. Our SPNet-P, SPNet-V and SPNet-PV improve the baseline by a large margin, with 1.36\%/1.34\%/1.08\% on the moderate level and 1.74\%/2.70\%/0.11\% on the hard level. The significant performance gains on the hard samples illustrate the importance of exploiting context information for detecting 3D objects with only a few points. Our approach SPNet-PV achieves the best performance among all competitors, pushing the moderate car AP to 84.92\%. The observations show that the AP gains on the KITTI test set are more significant than on the validation set. We argue that this results from the smaller domain gap between the train and validation set, which further illustrates the better generalization of our method.

\noindent\textbf{Results on “Pedestrian” and “Cyclist”} We also report the experimental results on small objects like pedestrians and cyclists, shown in Table \ref{kitti_ped_cyc}. Following previous works \cite{shi2019pointrcnn,zhou2018voxelnet}, the 3D detection results are evaluated IoU = 0.5 with 11 recall points. We adopt PointPillars \cite{lang2019pointpillars} as the baseline and improve the detection accuracy of pedestrians and cyclists from 54.98\%/65.19\% moderate AP to 56.46\%/68.03\% on the moderate setting, respectively. Especially for the cyclist class, our SPNet-P achieves significantly better performance than all other methods with up to 1.36\% and 1.08\% 3D AP improvement on the moderate and hard levels.

\subsection{Ablation Study}
We conduct extensive ablation studies to comprehend the roles of different components in SPNet. All experiments in this section are conducted on SPNet-P, which utilizes PointPillars \cite{lang2019pointpillars} as the baseline. We mainly focus on the car category and use IOU = 0.7 with 11 recall points as the evaluation metric.

\noindent\textbf{The effects of different components} The SPNet consists of three standalone information passing modules: \distillmethodone{}, \distillmethodtwo{} and \distillmethodthr{}. From the results shown in Table \ref{ablation}, we can find the final performance has been improved with the addition of any information passing module, with 0.92\%/1.16\%/0.43\% on the moderate setting, respectively. Adding the three components together further boosts performance, proving that our modules can complement each other. 

\begin{table}[t]
 \caption{Evaluation on the loss design of \distillmethodone{} module. The results are for the car class at IOU=0.7 (R11).}
 \centering
 \resizebox{\columnwidth}{!}{
 \begin{tabular}{c|ccc|ccc}
  \toprule
  \multirow{2}{*}{Method}                 & \multicolumn{3}{c|}{3D Detection(\%)} & \multicolumn{3}{c}{BEV Detection(\%)}   \\
                                           & Mod.           & Easy           & Hard     & Mod.           & Easy           & Hard       \\
\hline
  Baseline & 77.31 & 87.29 & 75.55 & 87.60 & 89.82 & 85.71 \\
  Direct transfer & 78.38 & 88.48 & 77.08 & 88.07 & 90.38 & 86.04 \\
  Affine map  & 78.57 & \textbf{88.80} & 77.22 & 88.25 & 90.23 & \textbf{86.64} \\
  Class cluster  & \textbf{78.67} & 88.71 & \textbf{77.29} & \textbf{88.29} & \textbf{90.42} & 86.41 \\
  \bottomrule
 \end{tabular}}
 \label{class-wise}
\end{table}
\begin{table}[t]
 \caption{Evaluation on the loss design of \distillmethodtwo{} module. The results are for the car class at IOU=0.7 (R11).}
 \centering
 \resizebox{\columnwidth}{!}{
 \begin{tabular}{c|ccc|ccc}
  \toprule
  Method                  & \multicolumn{3}{c|}{3D Detection(\%)} & \multicolumn{3}{c}{BEV Detection(\%)}   \\
                                           & Mod.           & Easy           & Hard     & Mod.           & Easy           & Hard       \\
\hline
Baseline & 77.31 & 87.29 & 75.55 & 87.60 & 89.82 & 85.71 \\
  L1 loss & 78.54 & 88.53 & 77.15 & 88.24 & 90.36 & \textbf{86.73} \\
  KLD loss  & 78.64 & \textbf{88.74} & 77.25 & 88.26 & 90.37 & \textbf{86.73} \\
  L2 loss & \textbf{78.67} & 88.71 & \textbf{77.29} & \textbf{88.29} & \textbf{90.42} & 86.41 \\
  \bottomrule
 \end{tabular}}
 \label{feature-wise}
\end{table}
\begin{table}[t]
 \caption{Evaluation on the loss design of \distillmethodthr{} module. The results are for the car class at IOU=0.7 (R11).}
 \centering
 \resizebox{\columnwidth}{!}{
 \begin{tabular}{c|ccc|ccc}
  \toprule
  Method                  & \multicolumn{3}{c|}{3D Detection(\%)} & \multicolumn{3}{c}{BEV Detection(\%)}   \\
                                           & Mod.           & Easy           & Hard     & Mod.           & Easy           & Hard       \\
\hline
Baseline & 77.31 & 87.29 & 75.55 & 87.60 & 89.82 & 85.71 \\
  L1 loss & 78.67 & 88.63 & 77.22 & 88.23 & 90.41 & \textbf{86.95} \\
  L2 loss  & 78.58 & 88.04 & 77.20 & 88.26 & 90.19 & 86.88 \\
  KLD loss & \textbf{78.67} & \textbf{88.71} & \textbf{77.29} & \textbf{88.29} & \textbf{90.42} & 86.41 \\
  \bottomrule
 \end{tabular}}
 \label{pred-wise}
\end{table}

\begin{figure}[t]
  \includegraphics[width=0.45\textwidth]{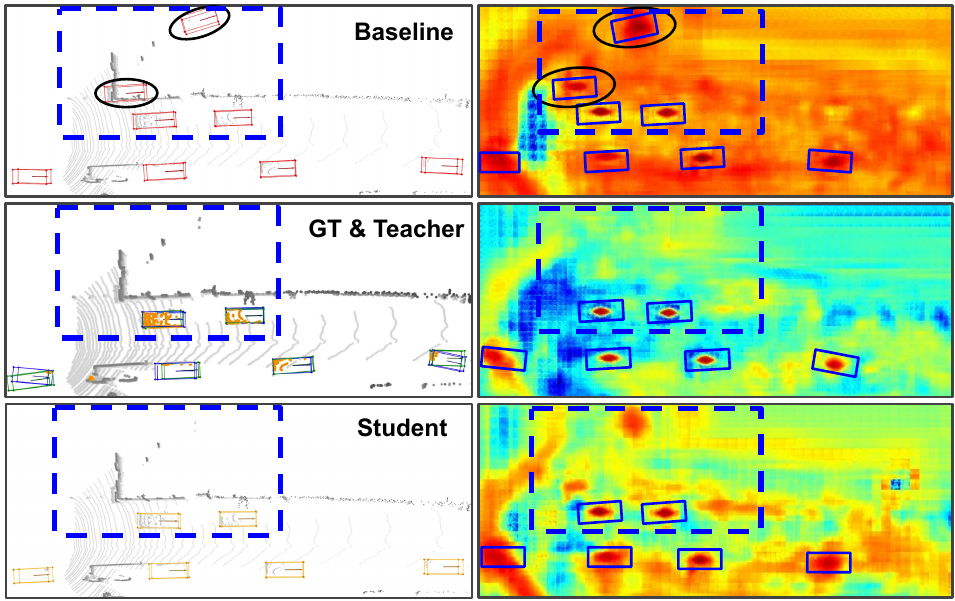}
 \caption{
Visualization of the 3D detection results and corresponding 2D feature maps at prediction level on KITTI validation split set. The predicted 3D boxes of the baseline model, the teacher and the student models are shown in red, green and orange, respectively, while the ground-truth boxes are in blue. The blue dashed boxes highlight the regions produce less false-positive and more accurate box predictions after applying our SPNet during training.}
 \label{obj_vis}
\end{figure}

\noindent\textbf{\distillmethodonecap{} module} We investigate different distillation strategies used in \distillmethodone{} module in Table \ref{class-wise}. We define some variants of the modules based on different styles of map generation: (a) Direct transfer: directly computing L2 loss between the feature maps; (b) Affine map: compute the affinity map of the feature maps from teacher and student model and calculate the L2 loss between the affinity maps; (c) Class cluster: our design.  Experimental results show that Direct transfer can improve the detection performance over the baseline, with 1.24\%/1.23\%/1.60\% improvement in the 3D detection. At the same time, the Affine map brings 1.26\% improvement over the baseline on the moderate level. Our Class cluster achieves the best result both on 3D detection and BEV detection, which demonstrates the effectiveness of the module.

\noindent\textbf{\distillmethodtwocap{} module} Table \ref{feature-wise} illustrates the impact of different losses used in the \distillmethodtwo{} module on the performance of the proposed model. We find that utilizing L1 loss, KLD loss and L2 loss can improve the baseline from 77.31\% to 78.54\%, 78.64\% and 78.67\%, respectively. Three losses achieve comparable results, and we choose the best L2 loss as the distillation loss for the \distillmethodtwo{} module.

\noindent\textbf{\distillmethodthrcap{} module} Table \ref{pred-wise} illustrates the impact of different losses used in the \distillmethodthr{} module on the performance of the proposed model. Similar to the \distillmethodtwo{} module, adding one of the L1 loss, L2 loss, or KLD loss can boost the performance of the baselines, and the improvements are comparable. However, we find that utilizing KLD loss here achieves the best performance. We claim that this results from the feature gaps between the middle and final output levels, and Kullback-Leibler divergence is more suitable for calculating the probability distribution distance for output instances. So we choose KLD loss to guide the learning of \distillmethodthr{} module.

\begin{table}[t]
 \caption{Evaluation on the encoding method design in the GT-Painting. The results are for the car class at IOU=0.7 (R11).}
 \centering
 \resizebox{\columnwidth}{!}{
 \begin{tabular}{c|ccc|ccc}
  \toprule
  Method                  & \multicolumn{3}{c|}{3D Detection(\%)} & \multicolumn{3}{c}{BEV Detection(\%)}   \\
                                           & Mod.           & Easy           & Hard     & Mod.           & Easy           & Hard       \\
\hline
Baseline & 77.31 & 87.29 & 75.55 & 87.60 & 89.82 & 85.71 \\
One-hot Encoding  & 87.13 & \textbf{89.94} & \textbf{80.04} & \textbf{90.21} & \textbf{90.74} & 90.16 \\
  Categorical Encoding & \textbf{87.42} & 89.76 & 79.98 & 90.19 & 90.62 & \textbf{90.20} \\
  \bottomrule
 \end{tabular}}
 \label{encoding_table}
\end{table}


\noindent\textbf{Encoding method for GT-Painting} Table \ref{encoding_table} illustrates the impact of using different encoding methods for GT-Painting. We define two variants: (a) One-hot Encoding: the class information is encoded into a one-hot vector with the shape of class numbers. For example, if a point belongs to a car, it will be appended with $[0,1,0,0]$. (b) Categorical Encoding: we use a scalar to indicate which category the point belongs to. The experiments show that both methods can improve the baseline by a large margin and comparable improvements. This demonstrates the importance of class-wise information whose contribution is not limited by the representative form. In order to reduce the computational complexity of the network, we choose Categorical Encoding in this paper.

\noindent\textbf{Visualization} Qualitative results are provided in Figure \ref{obj_vis}. We visualize the 3D objection detection results and the corresponding feature maps of three types of models: baseline, teacher and student, where we use PointPillars \cite{lang2019pointpillars} as the baseline here. Since the feature maps of the resulting space can better reflect the discrepancy between different models, we only present the feature maps taken from the instance-wise passing module. Obviously, the baseline will produce many false predictions, which illustrates that merely applying supervision in the label space can not fully exploit the semantic information contained in the objects. On the contrary, the teacher model can take advantage of the input semantics and produce more accurate results, then pass the semantic information to the student, achieving better results. The differences in the feature map further reflect this problem and simultaneously illustrate the necessity of our transfer semantics.

\section{Conclusion}
In this paper, we propose a novel Semantic Passing Network (SPNet) for 3D object detection. Our method can take full of the semantic information in the point cloud labels and distill the instructive knowledge to the student network during training. Benefited from this design, we can improve the existing 3D detectors by a large margin without any inference computational cost. Note the proposed SPNet achieves state-of-the-art 3D detection performance on the KITTI benchmark, which demonstrates the effectiveness of the proposed method.


\noindent\textbf{Limitation} So far, our work only considers the distillation point in the one-stage network or the first stage of the two-stage network, ignoring the second refinement stage. We plan to explore the influence of designing distillation losses in the second stage on the final performance for future work. Besides, we will apply our SPNet to more 3D object detectors to boost the localization performance.

\bibliographystyle{ACM-Reference-Format}
\balance
\bibliography{sample-base}

\appendix

\end{document}